\documentclass[10pt,journal,compsoc]{IEEEtran}

\usepackage[numbers]{natbib}

\usepackage{graphicx}
\usepackage{mathptmx}
\usepackage{graphicx}
\usepackage{amsmath,amssymb}
\usepackage{subfigure}
\usepackage{algorithm2e}
\usepackage{listings}

\begin{document}

\title{Data Survivability in Networks of Mobile Robots in Urban Disaster Environments}

\author{
Nicolas Kourtellis* \and
Adriana Iamnitchi* \and
Cristian Borcea$\dagger$ \and
Robin Murphy$\ddagger$ \\
*Department of Comp. Science \& Engineering, University of South Florida, Tampa, FL, USA\\
$\dagger$Department of Comp. Science, New Jersey Institute of Technology, University Heights, Newark, NJ, USA\\
$\ddagger$Department of Comp. Science and Engineering, Texas A\&M University, College Station, TX, USA\\
\{nkourtel@mail.usf.edu, anda@cse.usf.edu, borcea@cs.njit.edu, murphy@cse.tamu.edu\}
}


\date{}

\IEEEcompsoctitleabstractindextext{%
\begin{abstract}
Mobile multi-robot teams deployed for monitoring or search-and-rescue missions in urban disaster areas can greatly improve the quality of vital data collected on-site.
Analysis of such data can identify hazards and save lives.
Unfortunately, such real deployments at scale are cost prohibitive and robot failures lead to data loss.
Moreover, scaled-down deployments do not capture significant levels of interaction and communication complexity.
To tackle this problem, we propose novel mobility and failure generation frameworks that allow realistic simulations of mobile robot networks for large scale disaster scenarios.
Furthermore, since data replication techniques can improve the survivability of data collected during the operation, we propose an adaptive, scalable data replication technique that achieves high data survivability with low overhead.
Our technique considers the anticipated robot failures and robot heterogeneity to decide how aggressively to replicate data.
In addition, it considers survivability priorities, with some data requiring more effort to be saved than others. 
Using our novel simulation generation frameworks, we compare our adaptive technique with flooding and broadcast-based replication techniques and show that for failure rates of up to 60\% it ensures better data survivability with lower communication costs.
\end{abstract}

}

\maketitle

\section{Introduction}
\label{sec:intro}

Recent technology advancements make robots valuable partners in real-life mission-critical scenarios.
Robots proved crucial in disaster and rescue missions such as terrorist attacks on civilians~\cite{murphy04security, murphy01wtc}, natural disasters such as hurricanes~\cite{murphy11ike, murphy05katrina1} and earthquakes~\cite{spectrum11japansurvivors, spectrum11japanreactors2}, and life-threatening mining accidents~\cite{murphy08mine}.
However, despite these advances, there is still a significant gap between laboratory work and real-life environments, as acknowledged by various robotic competitions~\cite{robotnet11competition} and on-going field research in this area~\cite{theeagle11robotstorescue}.

In ad-hoc collaborating robot teams this gap is highly noticeable.
Current robust deployments are typically done at a small scale and using teleoperation from human coordinators for reconnaissance and mapping of a disaster area, assessment of the damages and identification of dangerous zones (e.g.,~\cite{birk11mosaicking, kleiner11mapping, murphy11ike}).
Robots in such teams typically do not communicate with each other, but collect and transmit information either wirelessly or through tethering to a base station.

Deployments where robots are more autonomous and collaborate with each other to achieve a global task in critical missions~\cite{maza10multi-uav} have not yet proven to scale to hundreds or thousands of robots.
Thus, the full potential of large teams of collaborating robots collecting crucial information in time-pressure, life-threatening, real-life scenarios is yet to be realized.
For example, such a scenario could involve thousands of micro-robots (e.g., inch-sized robots~\cite{dudenhoeffer00micro-robot-formation}, palm-sized helicopters~\cite{bouabdallah10palm-sizeheli} or quadrocopters~\cite{burkle10uav-swarms}) or even fly-sized robots~\cite{wood08flyrobot}, to collect data on human survivors in a cataclysm.
At this scale, humans can be only partially involved in controlling them; instead, the robots should reliably collaborate and coordinate with each other and the environment in an ad-hoc manner.

A major obstacle that delays the appearance of such large-scale self-coordinating mobile robot networks is the reliability of the robots in the presence of environmental hazards.
High failure rates lead to significant loss of data.
Data replication can be used to improve the overall data survivability.
However, existing techniques proposed in literature, and especially for mobile ad-hoc networks (MANETs), do not work well in mobile robot networks targeting disaster scenarios for three reasons.
First, these solutions assume homogeneous nodes, while robot teams may be heterogeneous, with different capabilities and failure rates.
Second, robot failures are not always independent as it is assumed in these methods.
Third, the existing techniques have been designed for much denser networks that typically ensure network connectivity between any two nodes.
Robot networks, on the other hand, are sparse, and reachability between nodes cannot be guaranteed, especially in urban-disaster scenarios.
Thus, sending all the collected data to a base station in real-time is not possible.

To tackle the problem of data survivability under realistic mobile robot network conditions, we propose an adaptive, scalable data replication technique that takes into account the neighborhood robot density and the rate and type of robot failures to decide how aggressively to replicate data for higher survivability.
This technique is based on opportunistic communication, similar to delay tolerant networking, to maintain system flexibility, reliability and robustness in face of robot and network failures~\cite{frew09networking}.
In addition, inspired from real-world scenarios, our technique includes a new parameter that represents the survivability requirements of data produced by robots.
This parameter reflects the importance of different data types to the human crews and determines their replication needs during the mission.

We also propose novel frameworks for generating realistic mobility and failure scenarios for teams of robots working in urban disaster environments, based on a thorough analysis of prototype robots in real-life monitoring situations~\cite{carson03reliability, carson05failures, carson04failures}.
The mobility generation framework allows variable type, number, and size of deployment areas, variable type and number of robots deployed in each area, and different type of mobility for each area.
Furthermore, the failure generation framework realizes three types of failures: independent robot failures, specific to technical malfunctions; independent area failures, specific to problems such as collapsed bridges or buildings; and clustered areas failures, specific to major disasters such as explosions or broken levees.
Via simulations using these frameworks, and under extreme failure rates, we demonstrate that our adaptive replication technique results in higher overall data survivability than baseline techniques for comparable communication costs.

The rest of this paper is organized as follows:
Section~\ref{sec:scenario} describes how large-scale mobile robot teams could work in real-life situations.
Section~\ref{sec:robot-nets} presents the design of our mobility generation framework when considering real-life scenarios of robotic missions in large operational areas where robots are assigned roles and areas to operate in.
Section~\ref{sec:failure-model} defines analytically the failure generation framework considered in this study and proposes three failure models applicable to various real mission scenarios.
Section~\ref{sec:repl-surv} discusses replication techniques for data survivability.
We evaluate and compare the replication techniques using our mobility and failure generation frameworks via extensive simulations in Section~\ref{sec:evaluation}.
Section~\ref{sec:rel-work} discusses related work, and Section~\ref{sec:conclusions} elaborates on our experimental findings and concludes this study.

\section{Mobile Robot Team in an Operational Scenario}
\label{sec:scenario}

\begin{figure*}[htbp]
  \centering
	\includegraphics[scale=0.9]{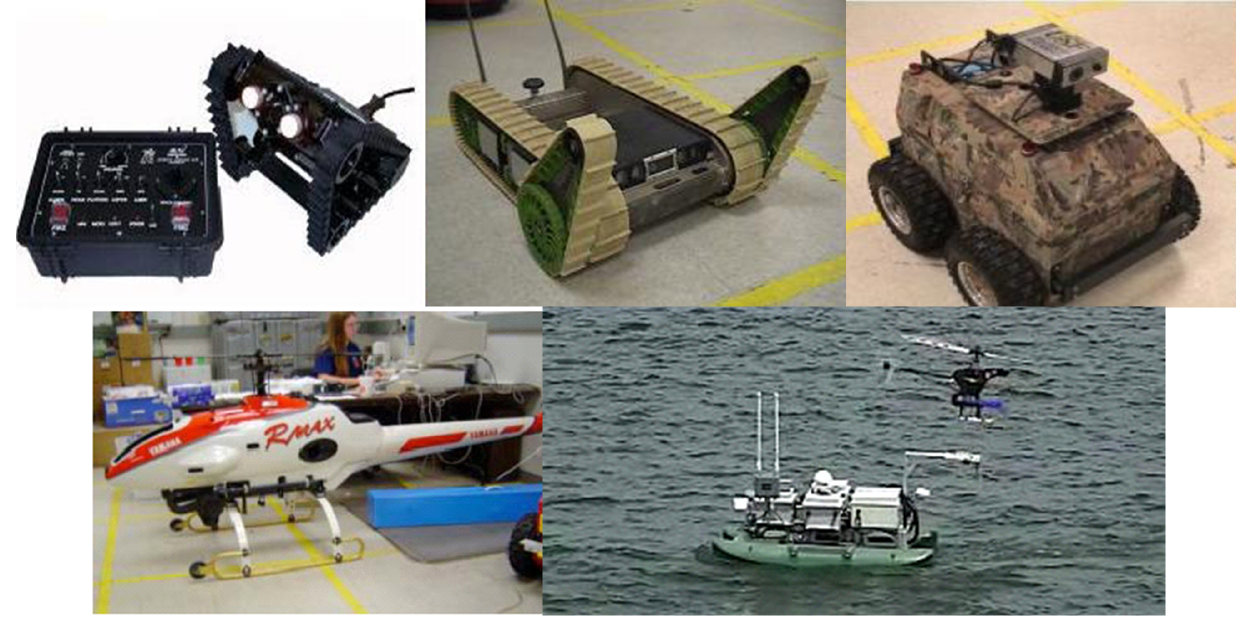}
      \caption{Examples of unmanned (ground, aerial and surface) robotic vehicles used in human-robot missions}\label{fig:crasar-robots}
\end{figure*}

Today, various types of robots are used in search and rescue missions and other human-robot tasks.
These robots are unmanned vehicles with an autonomic operation of a few hours, depending on the usage of the on-board components.
A typical robot, such as the ones operated by CRASAR~\cite{crasar11crasar} and shown in Figure~\ref{fig:crasar-robots}, has a powerful system on-board, usually close to today's PCs.
Along with the computation platform, other hardware modules installed may include GPS, inertial measurement unit, laser, cameras, temperature sensors, and network cards.

\begin{figure*}[htbp]
	\centering
	\includegraphics[scale=1.2]{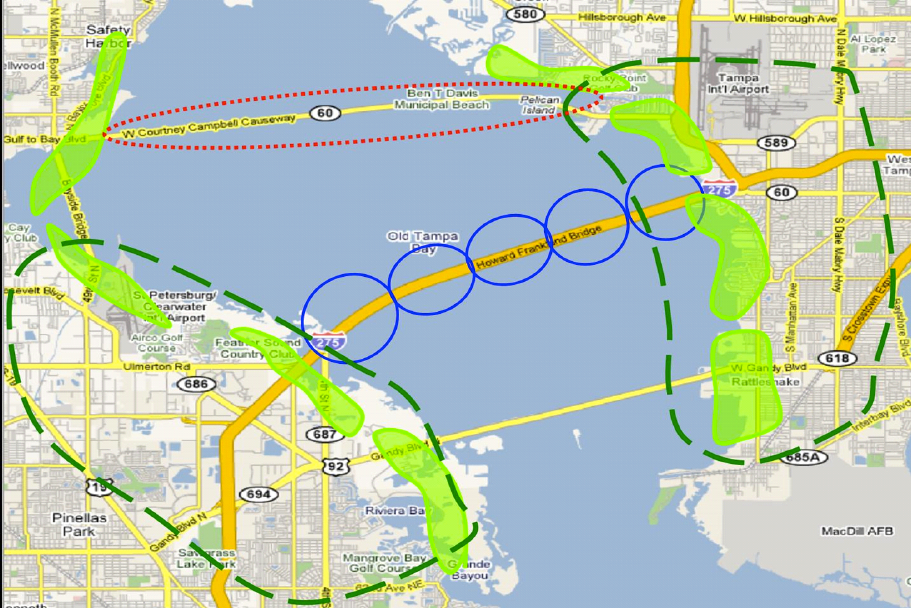}
	\caption{Operational scenario of a multi-robot littoral deployment.
	Continuous circles represent small areas of the bridge scanned by UAVs and USVs.
	The dotted elliptical line gives the flight path of a low altitude UAV monitoring the small teams assigned to each bridge.
	Light shaded areas represent swarms of UAVs, whereas dark long dashed lines show the coverage offered by higher altitude UAVs.}
	\label{fig:scenario}
\end{figure*}

A potential operating scenario for a team of robots is presented in Figure~\ref{fig:scenario}.
After a strong hurricane, much of the civilian infrastructure (bridges, roads, and buildings) has been affected.
The deployment of human personnel for assessing the damage is considered extremely dangerous, and the authorities decide to deploy a team of robots of various types, such as unmanned ground vehicles (UGVs), unmanned aerial vehicles (UAVs), and unmanned surface vehicles (USVs).
The three bridges need to be inspected to assess the magnitude of the damage.
This scenario emphasizes the distributed interaction between robots and identifies the parameters that need to be modeled for these networks.

Three ``fleets'' of USVs enter the bay and each fleet is assigned the task to examine one bridge.
A fleet consists of a mother ship which deploys teams of robots comprised of an USV and two rotary wing low-altitude micro-aerial vehicles.
These teams are assigned to portions of each bridge.
The aerial vehicles provide support to the surface vehicles.
Each mother ship circles the sub-bay area of the bridge to provide support and delay-tolerant network connectivity.
In addition, there are a low-altitude fixed wing vehicle and a ground vehicle providing surveillance for each bridge.
Eight swarms of mid-altitude fixed wing aerial vehicles cover the coast.
Each swarm has five UAVs flying in a coverage pattern over a portion of the coastal area.
Two higher altitude aerial vehicles circle the remaining ground area to provide a total aerial coverage.
The bridges are also inspected by three convoys of ten UGVs each.

The teams produce data using their on-board sensors and use ad-hoc networking to coordinate with each other.
When the mission ends, they travel to a predefined location, where data are collected and analyzed in greater detail to assess the damage to each bridge.
Therefore, it is of critical importance that the data produced by each robot are stored in the system in a resilient manner to survive the mission.
Currently, the survivability of data is directly dependent on the fault tolerance of the robots and their intelligence to stay off dangerous paths or areas until the mission is over.
Unfortunately, in this type of disaster missions, robots fail frequently due to environmental conditions (e.g., explosions, collapsed buildings, etc.) as well as hardware problems.
Thus, ad-hoc networking and collaboration between robots must be used to increase data survivability.

The operational scenario described above helps us identify the following parameters to be modeled in our mobility and failure generation frameworks:
(1)~the scale of the operation area that all robots are situated in,
(2)~the types and sizes of sub-areas that particular robots are assigned to work in,
(3)~the types and number of robots available during the operation, acquiring different mission roles and assigned in particular types of sub-areas,
(4)~the types of mobility for particular types of robots,
(5)~the types and rates of failures anticipated by particular types of robots during the mission.
We explore these operational parameters in the next sections, where we describe in more detail the mobility and failure generation frameworks considered in this study.

\section{Modeling Mobility in Robot Networks}
\label{sec:robot-nets}

Because large-scale real deployments are cost prohibitive and scaled-down deployments do not capture a significant level of interaction and communication complexity for our study, simulations are necessary for understanding the behavior of mobile robot networks at large scales.
To realistically simulate a mobile robot network, we created a framework that models robot assignment to particular areas, and robot mobility according to their mission role.
This framework takes into consideration the following characteristics of mobile robot networks:

\begin{itemize}
\item Mobile robot networks are sparse, with node degrees of $2$ or $3$.
\item The robots are not uniformly distributed in the area of operation; instead, they are clustered in relatively small regions and assigned to work collaboratively on a task.
Large parts of the operation area may, at times, have no robot presence.
\item The operation area in a disaster scenario could spread over tens of squared kilometers (e.g., the scenario presented in Section~\ref{sec:scenario} covers about $25Km^2$).
\item Depending on their assigned task, robots could move systematically to cover all their working area when mapping or searching for a particular item or have a random-like mobility pattern when more greedy searching approaches are used.
\item Usually, end-to-end communication is not required between all pairs of robots because of localized collaboration.
Consequently, routing between robots of the same cluster is not a major problem since clusters that need routing have nodes with low mobility (e.g., $1$--$5m/s$).
\item Robots in such large deployments can be highly heterogeneous, with different capabilities, different failure rates, and different assignments and roles: in one scenario~\cite{sugiyama05collaboration}, for example, search robots scan a disaster area for victims, while relay robots propagate data.
More powerful nodes, such as UAVs and USVs can act as coordinators or data collectors for the others (e.g., UGVs).
\end{itemize}

\begin{figure*}[htbp]
  \centering
	\includegraphics[scale=0.9]{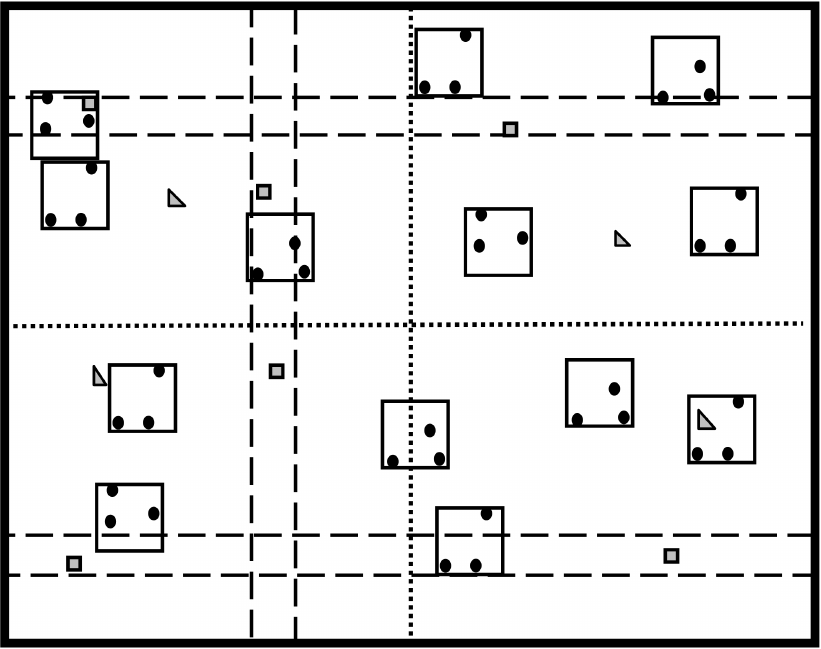}
      \caption{Example of generated mobility scenario.
      The boxes represent different types of areas where robots are assigned and allowed to move.
      The robots stay in their areas, but interact with their group-mates and other areas' robots.
      Small squares with continuous lines are working areas, stripes (vertical or horizontal) with dashed lines are connecting areas, and dotted lines are monitoring areas.
      Black circles represent scouts, grey squares represent archivists, and grey triangles represent supervisors.
      }\label{fig:network-scenario}
\end{figure*}

Based on these characteristics, we identified the following parameters for modeling mobile robot networks: (1) area type, (2) area size, (3) robot types per area, (4) number of robots per area, and (5) type of mobility per area.
Figure~\ref{fig:network-scenario} illustrates a network scenario with three types of areas and the corresponding types of robots assigned to each.
The specifics of these types of areas and robots are discussed next.

\emph{\textbf{Area Types in Operation Area}}:
Typical tasks of robot networks include missions that cover large portions of terrain, water, or both.
Thus, the overall operation area can be as large as tens of squared kilometers.
In such an operation area, we define working, connecting and monitoring areas.
We note that the mobility, type and number of robots appointed to each of these areas are intrinsically connected to the mission roles assigned to the robots and the subtasks they are to perform.

\emph{Working Areas} are typically covered by small teams of collaborative robots that perform various subtasks under a particular role like \textit{scouting}.
These robots, \textit{scouts}, move slowly ($1$--$5m/s$), acquiring and analyzing data, while inspecting an area (e.g., checking the structural integrity of bridge legs and damaged buildings), and searching for victims.
They react and adapt to the current context and collaborate in groups to achieve a global task.
Since scouts are assigned to areas expected to fail, valuable data collected by their on-board sensors must be replicated in order to increase their survivability.
The working area size depends on the mission, but typically is about the size of a building (e.g., $100m \times 100m$).
These areas are either known before the mission starts or determined dynamically as a function of the changing context in the field.
In our framework, we consider that working areas do not overlap, but robots from neighboring regions can communicate with each other if they are in transmission range.

\emph{Connecting Areas}: The operation area could be very large and scouts in different working regions could form multiple disconnected network partitions.
More powerful robots cover connecting areas, moving across or around the working areas, interacting with or controlling the scouts.
These robots are given the role of \textit{archivist}.
The archivists can be large vehicles that move slowly enough to manage their way through potential obstacles ($5$--$10m/s$).
They collect and log data from scouts in on-board resilient storage until the end of the mission and connect the network partitions in a delay tolerant fashion, providing support for data mulling~\cite{vasilescu05muling}.
In an urban scenario with a grid road layout, the connecting areas could be avenues and streets.
In general, they can be long stripes covering the whole length or width of the operation area.
Similar to the working regions, they can be known a priori or assigned dynamically as a function of the position of the working regions and the paths taken by the human crews.

\emph{Monitoring Areas}: UAVs flying at fairly low altitudes and at speeds up to $20$--$30m/s$ can offer a high-level monitoring of the operation area and alert the human crews in case of an emergency.
Several monitoring regions can cover the entire operation area, but do not need to overlap.
During their pass over the working regions, these UAVs can also log data from scouts on resilient storage until the mission ends, realizing the combined roles of \textit{supervisor} and \textit{archivist}.

\emph{\textbf{Robot Mobility}}:
Our framework allows specifying different mobility models for individual robots or groups of robots, depending on the heterogeneity of the network. In choosing the mobility pattern for robots in each area, we must consider the mission type and the robot roles. For example, a supervisor could be instructed to move with a certain average speed between all working areas within its monitoring area, and slow down when flying over them. An archivist could move through its assigned area and slow down when communication with scouts is necessary. Finally, for scouts in a small working area, the random waypoint mobility could work just fine. More complex group mobility or mobility driven by the mission's context can be considered as well.

\section{Modeling Failures in Robot Networks}
\label{sec:failure-model}

Node failures in robotic networks or MANETs are usually considered independent (if failures are considered at all).
Only few studies~\cite{kotsovinos05replic8} have considered clustered node failures that affect multiple nodes at the same time.
In addition to the commonly examined scenario of isolated technical problems, such as motor issues (e.g., due to small rocks, ponds, vegetation or sandpits) or running out of battery, in this study we consider two clustered node failure scenarios that we model next.
First, isolated explosions or building collapses can affect multiple robots working in the same area.
Second, unstable environmental conditions such as a big gas explosion or a bridge collapsing can cause failures to multiple such areas and their assigned robots.
Furthermore, for each of these failure scenarios there is an associated probability of a robot's neighborhood to fail.
In Section~\ref{fp} we present analytical models that enable a robot to estimate this probability and adjust its data replication accordingly (used in Section~\ref{sec:repl-surv}).

\subsection{Failure Models}

We propose a novel failure generation framework, that implements three types of failures: independent robot failures, independent area failures, and clustered area failures.
In this framework we consider permanent failures of robots: when a failure happens, the robot stops moving and communicating and the on-board collected data are considered irretrievable.

\emph{\textbf{Model 1: Independent robot failures.}} This model covers the most frequently occurring scenario where robots fail independently of each other.
In this model, each robot type has an associated failure rate.
Usually, scouts experience higher failure rates because of their assigned tasks (working areas), while the archivists and supervisors are the most reliable due to their powerful resources and less cluttered paths.
We consider these failures are uniformly distributed over the operation area and the duration of the mission, as a function of a pre-established robot failure rate based on the particular type of robot and task assigned.

\emph{\textbf{Model 2: Independent area failures.}} This model covers the real-life scenarios where all robots in a working area fail altogether (e.g., due to a building collapse).
We consider these failures are uniformly distributed over the working areas and duration of the mission, using a pre-defined area failure rate.
Thus, during the operation, working areas are randomly selected to fail and all robots assigned to such areas fail simultaneously.

\emph{\textbf{Model 3: Clustered area failures.}} This model covers the case where an event affects multiple neighboring working areas, failing all their containing robots: a big explosion, a land slide, or a broken levee could lead to subsequent or simultaneous failures in neighboring areas.
We use a pre-defined failure rate to define the total number of failing neighboring working areas to account for the scale of the event.
Thus, during the operation, a random working area $A$ is selected to fail, and all working areas neighboring $A$ also fail simultaneously.
The number of neighboring areas to fail depends on the pre-defined failure rate.
All robots assigned to these clustered areas fail simultaneously.
The difference between independent area failures and clustered area failures of the same failure rate is in the location of failing robots: in scattered, randomly selected areas in the former case, or all concentrated in neighboring areas in the latter case.
The two models have different outcomes in terms of network connectivity/partitioning and, as shown by our experimental work, in the performance of data replication algorithms.

The above models assume that during a particular mission, one of the three types of failures dominates the operation.
The human crews can assess which model to use given the environmental conditions and the type of disaster under investigation.
Also, they can over-estimate the expected failure rate and reduce it to lower levels when they have updated information on the mission's status.

We can also combine the above failure types to produce a more complex model based on a mixture of failure rates for each of the models, where robots can fail independently of each other (as in the first model), in small independent groups (as in the second model) or in large clustered groups (as in the third model).
In this work we examine individually each of the three models; in the future, we plan to explore their combination into a more complex model.
Next, given a particular failure model, we define analytically the combined probability of failure of a neighborhood of robots, as a function of the different type and number of robots or areas comprising this neighborhood and their expected failure rates.

\subsection{Failure Probability}\label{fp}

Data collected by the on-board sensors of a robot such as a scout can be lost due to robot failures.
Data replication to nearby robots, especially to more reliable types of robots such as archivists or supervisors, can increase the survivability of these data.
During a mission, if a robot can assess the probability of its neighboring robots to fail, it could adjust how aggressively to replicate data.
Therefore, the \emph{Failure Probability ($FP$)} of the robot's neighborhood can be estimated given the expected failure model, and the failure rate and number of neighboring robots (first model) or failure rate and number of neighboring areas (second and third model).
$FP$, defined next for each failure model, implies that fewer robots or working areas in the neighborhood increases the probability for data loss, but it also depends on the robot or area types.
For example, if the neighboring robots are only scouts, the failure probability will be higher than if there is a neighboring archivist as well.

During a mission under failure model $m \in M =\{1, 2, 3\}$, we assume the following for the neighborhood of a robot:
\begin{itemize}
\item There are robots of different types $i \in RT = \{1, 2, 3, \dots\}$ (e.g., $i$=$1$ for scouts, $i$=$2$ for archivists, $i$=$3$ for supervisors, etc).
\item There are $a_i$ number of robots of a type $i \in RT$.
\item Each robot type has an anticipated failure rate $\rho_{i}$.
\item There are neighboring robot areas of different types $j \in AT = \{1, 2, 3, \dots\}$ (e.g., $j$=$1$ for working areas, $j$=$2$ for connecting areas, $j$=$3$ for monitoring areas, etc).
\item There are $b_j$ number of areas  of a type $j \in AT$.
\item Each robot area has an anticipated failure rate $\lambda_{j}$.
\end{itemize}

Then, we define the Failure Probability $FP_m$ for each failure model $m \in M$ as follows:

For \textit{independent robot failures} ($m=1$) and robot types $i \in RT$:
\begin{equation}\label{eq:1}
FP_1 = \rho_{1}^{a_1}, \text{ }(\rho_{1})^{a_1}(\rho_{2})^{a_2}, \text{ }\cdots, \text{ }\prod_{i=1}^{RT} (\rho_{i})^{a_i}.
\end{equation}

For \textit{independent area failures} ($m=2$) and robot area types $j \in AT$:
\begin{equation}\label{eq:2}
FP_2 = \lambda_{1}^{b_1}, \text{ }(\lambda_{1})^{b_1}(\lambda_{2})^{b_2}, \text{ }\cdots, \text{ }\prod_{j=1}^{AT} (\lambda_{j})^{b_j}.
\end{equation}

For \textit{clustered area failures} ($m=3$) and robot area types $j \in AT$:
\begin{equation}\label{eq:3}
FP_3 = \lambda_{1}, \text{ }(\lambda_{1})(\lambda_{2}), \text{ }\cdots, \text{ }\prod_{j=1}^{AT} (\lambda_{j}).
\end{equation}

As the three proposed models show, the calculation of $FP$ takes into account multiple types of robots and robot areas.
To demonstrate how the failure probability changes across different failure models and rates, we consider the following example, where we limit the types of robots to scouts (SC) and archivists (AR) and the types of areas to working (WA) and connecting areas (CA).
In this example, we assume that a robot calculating the $FP$ of its neighborhood encounters the following four typical scenarios of combinations of robots and areas:
\begin{enumerate}
\item Two scouts assigned to the same working area ($2SC$-$1WA$)
\item Two scouts assigned to two different working areas ($2SC$-$2WA$)
\item Two scouts assigned to the same working area, and one archivist assigned to a connecting area ($2SC$-$1AR$-$1WA$-$1CA$)
\item Two scouts assigned to two different working areas, and one archivist assigned to a connecting area ($2SC$-$1AR$-$2WA$-$1CA$)
\end{enumerate}

\begin{figure*}[htbp]
	\centering
	\includegraphics[scale=0.55]{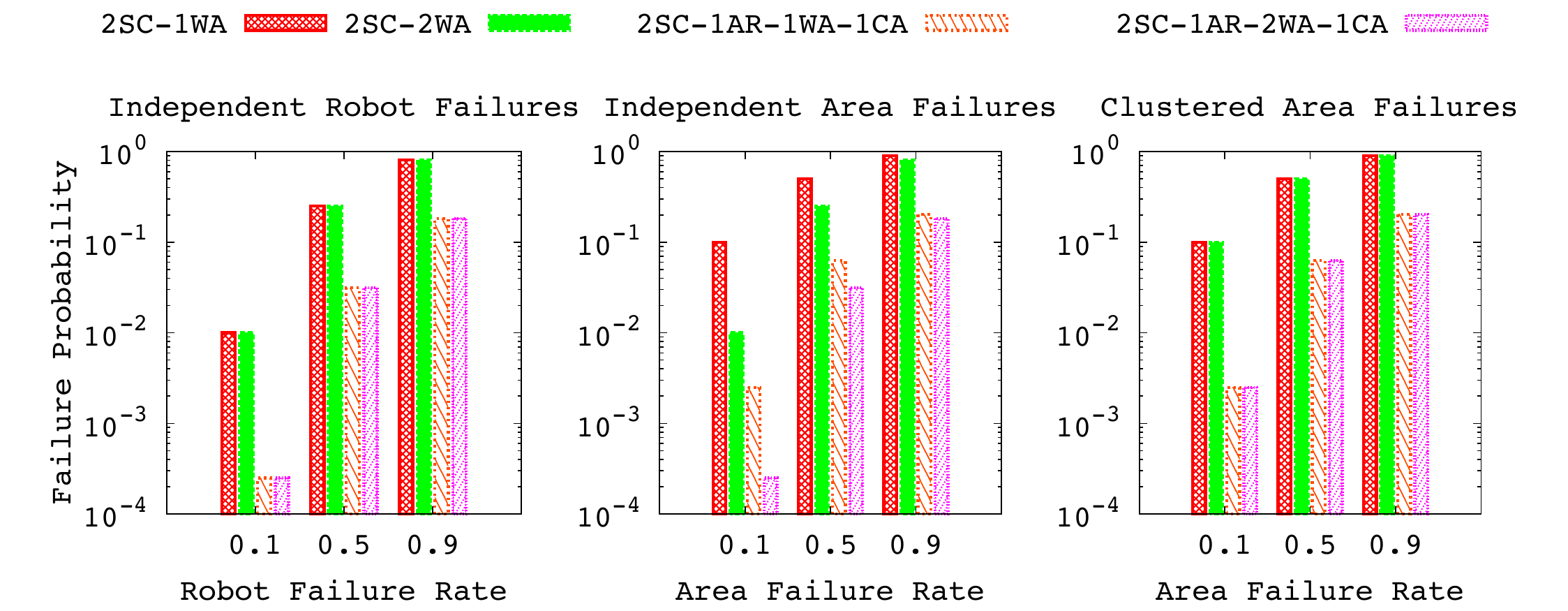}
	\caption{
	Failure probability with respect to failure rate of scouts (SC) and archivists (AR) when assigned to working (WA) and connecting areas (CA), under the three failure models and for four hypothetical scenarios.
	For the independent robot failure model, we assume that scouts have 4 times the failure rate of the archivists (i.e., $\rho_1=4\rho_2$).
	For the independent area and clustered area failure models, we assume that working areas have 4 times the failure rate of the connecting areas (i.e., $\lambda_1=4\lambda_2$).
	}
      \label{fig:failure-type-demo-small}
\end{figure*}

Figure~\ref{fig:failure-type-demo-small} illustrates the failure probability given a particular scenario, under the three failure models and three failure rates applied.
We set the failure rate of the archivists (and respectively of the connecting areas) to $1/4$ of the failure rate of the scouts (and respectively of the working areas), since we assume they exhibit higher reliability due to better hardware and safer task assigned.
The results verify our intuition that the lowest $FP$ for all scenarios and failure rates is anticipated at the independent robot failures.
Furthermore, the combination of robot types and areas assigned greatly affects the estimated failure probability.
For example, the presence of a single archivist in the neighborhood of the transmitting robot can lower the failure probability for all failure types and rates, in comparison to just having neighboring scouts.
Also, having neighboring scouts assigned to different working areas decreases the $FP$ when comparing independent robot and area failures.
The highest $FP$ is expected in the clustered area failures where multiple robots from neighboring areas can fail simultaneously.
In fact, in some scenarios there is a multifold increase of $FP$ between independent robot and area failures (e.g., $2SC$-$1WA$ and $2SC$-$1AR$-$1WA$-$1CA$ under a failure rate of $0.1$), or between independent area and clustered failures (e.g., $2SC$-$2WA$ and $2SC$-$1AR$-$2WA$-$1CA$ under a failure rate of $0.1$).

\section{Replication for Data Survivability}
\label{sec:repl-surv}

Data replication can increase the level of data survivability across the robot network when under robot failures.
In such disaster missions, the data collected may have different survivability requirements, depending on their importance to the human crews.
For example, infrared video footage inside a collapsed mine can be used to identify locations where people are trapped, thus could require high survivability.
Also, radiation level readings inside a nuclear reactor after a power plant accident~\cite{spectrum11japanreactors1} can help the emergency crews identify dangerous locations, or when it is safe to enter the facility and perform repairs.
However, humidity level readings might not be as important to survive the mission, therefore could require low survivability.
Thus, adapting replication to the type of data can make better use of network resources while better serving the requirements for data survivability.

This section discusses techniques for data replication for increased survivability with respect to cost of communication overhead, storage space, and power consumption.
Understanding the tradeoff between the degree of data replication and the network costs is the objective of our work.
To this end, we start our investigation with two intuitive techniques that do not take into account robot or data type in the data replication decision: (1) flooding, that optimizes uniform data survivability for all data at high network costs; and (2) broadcasting, that minimizes network costs with an intuitive penalty in data survivability of all types of data.
While these techniques are well understood and have been thoroughly investigated in the past, our experimental evaluations contribute new understandings for environments specific to robot teams.
We continue with two other techniques that combine broadcasting and flooding but also take into account robot heterogeneity based on type of robot and task assigned.
In particular, these techniques distinguish between scouts, mainly responsible to record data from the area of operation but prone to failures, and archivists, reliable nodes that ensure data delivery at the end of the mission.
Finally, we propose an adaptive technique in Section~\ref{subsec:adaptive} which takes into consideration not only the different roles (i.e. types) and anticipated failures of the robots, but also the different types of data collected and their different survivability requirements.

\subsection{Basic Data Replication Techniques: Flooding and Broadcasting}

\textbf{Flooding (Fl)}: A straightforward replication technique is flooding.
The robot-source transmits a message or file with a broadcast to all neighboring robots.
The robots within transmission range that receive this message, independently of each other, decrement the time-to-live ($TTL$) (number of hops in the robot network) and retransmit if $TTL>0$.
This process is repeated until $TTL$ becomes zero in all copies of the message.
Flooding with large $TTL$ achieves high data survivability in mobile ad-hoc networks due to higher connectivity.
However, a robot network is usually sparse and with varying connectivity based on the mobility model, weather conditions, terrain and other obstacles.
Thus, flooding even with a high $TTL$ can introduce redundant retransmissions and unnecessary usage of the robot wireless network interfaces and thus increased power consumption, without translating to increased data survivability, since the message may fail to reach nodes away from a failing area.
Our experimental results in Section~\ref{sec:evaluation} provide more information on this topic.

\textbf{Broadcasting (Br)}: Another simple data replication technique is to just broadcast the data as soon as they are created.
All robots in transmission range receive and store locally the data broadcasted.
The replication, hence, stops at this first hop, and these data are never replicated again.
This is a special case of the flooding technique, with $TTL=1$.
This approach reduces network overhead and congestion in comparison to flooding, but it is likely to replicate less and thus lead to lower data survivability with respect to robot failures.

\subsection{Combined Flooding-Broadcasting Proactive Techniques for Data Replication}

To exploit the heterogeneity of robots deployed in a mission, we consider the case where archivists and supervisors broadcast regularly $HELLO$ messages while moving in their designated areas.
These messages alert the scouts in proximity to send cumulatively all data they produced since the last received $HELLO$ message.
This action increases the probability of data to survive a failure of the producer and its replicas on the scouts nearby.
The cumulative transmissions, however, subject the network to congestion: the data bursts transmitted by all scouts around an archivist in response to its $HELLO$ message lead to the broadcast storm problem~\cite{sze-yao99broadcaststorm}.

\textbf{Broadcasting and Cumulative Limited Flooding (BrCLFl)}: In this approach, the scouts broadcast their data regularly.
When they receive a $HELLO$ message from an archivist, they cumulatively flood all their data since the last received $HELLO$ message, with a limited $TTL$ (much smaller than the full flooding).
This technique tries to ensure that the data will reach the archivist within a few hops, even if the archivist was borderline passing the working area of the scout.
An increased network overhead is expected, yet lower than that of the simple flooding approach with high $TTL$.
However, simultaneous and cumulative data transmissions can lead to congestion at the archivist's wireless network interface and might limit data survivability.

\textbf{Broadcasting and Cumulative Broadcasting (BrCBr)}: Instead of flooding, the scouts cumulatively rebroadcast all their data since the last received $HELLO$ message.
This technique greatly reduces the network overhead, while trying to reach the archivist and increase the data survivability subject to network congestion and archivist reachability within the first network hop from the scouts.

\subsection{Failure-Adaptive and Delay-Tolerant Data Replication}\label{subsec:adaptive}

The above techniques explore the trade-off between data survivability and network overhead imposed by replication without fully exploiting the particularities of the robot environment:
\begin{enumerate}
\item Data have various degrees of importance and can be tagged accordingly with a desired level of survivability;
\item Robot failures can be anticipated based on the type of robots and knowledge about the operation area. If failures are common, data need more replication.
\end{enumerate}
We propose a scalable technique that allows each scout to adapt the replication level by taking into account the anticipated failure rate of its $1$--hop network neighborhood in conjunction with the survivability requirement of the data type to be replicated and the replication level it acquired from previous transmissions.
We assume that before the mission, the human coordinators assess the situation and decide what type of failure model is expected and give an estimate of the failure rate.
This failure rate can be set to a high level at the beginning of the operation, reflecting a pessimistic assessment of the situation and gradually change to lower levels as the coordinators have more empirical evidence at hand.
They can also define a priority across different types of data which can be used to assess the survivability requirement for each data type.
For example, in an earthquake scenario such as in Haiti~\cite{haiti10earthquake} or Japan~\cite{japan11earthquake}, infrared video can be assigned survivability requirement of 100\% and humidity or temperature readings a survivability requirement of 50\%.
All robots transmit regularly $HELLO$ messages that include their assigned area, expected failure rate, and data repository index.

Algorithm~\ref{algo:pseudo-code} shows the high-level pseudo-code for the adaptive replication technique.
When a robot is about to send a data item (line $5$), the Failure Probability $FP$ of the $1$--hop network neighborhood is assessed (line $7$) using the appropriate equation for the assumed failure model (i.e., eq.~\ref{eq:1}, eq.~\ref{eq:2} or eq.~\ref{eq:3}).
This is possible due to the $HELLO$ messages exchanged every few seconds between the robots.
$FP$ is an estimate of the survivability to be lost by the data item when broadcasted at that time in the $1$--hop neighborhood (as explained in Section~\ref{sec:failure-model}).
Thus, $1-FP$ is the survivability acquired from the neighborhood.
The remainder of the survivability for the item, from its originally set value, is calculated in line $8$.
If the number of neighbor robots that do not have the item (line $8$, as assessed by the $HELLO$ messages) and the survivability remainder (line $10$) are positive, the item's updated survivability requirement from the neighboring robots is defined as in line $11$.
Otherwise, it is set to zero because no more replication will be needed (line $14$).
In either case, this updated value is attached to the item, and the item enters the queue for immediate transmissions (line $16$) from the particular robot.
If there are no robots in the neighborhood or all of them have the specific item already stored (line $18$), it means that $FP$ was calculated taking into account only the local robot failure rate in line $7$.
In this case, the item enters the queue for delayed  transmissions (lines $19$--$20$).
The delayed transmissions queue $Q$ is examined every time the robot receives a new message such as a $HELLO$ (line $38$) or data item from other scouts (line $46$).
If $Q$ is not empty, each delayed data item is examined for any remainder of survivability requirement (line $31$) and the $ScoutSendData()$ function is called (line $32$) to handle it accordingly.
A scout receiving an item (line $40$), stores it locally (line $42$) and if there is survivability remainder, the $ScoutSendData()$ function is called (line $44$).
Otherwise, if an archivist receives an item (line $48$), it just stores it locally (line $50$).

\LinesNumbered
\begin{algorithm}
		$B = \{\};$ // Data Queue for immediate transmissions\\
		$Q = \{\};$ // Data Queue for delayed transmissions\\
		$S = \{\};$ // Local Data Storage\\
		$failure\_type \in \{1, 2, 3\}$\\
		$\textbf{\emph{ScoutSendData}}(data\{dataID,surv\_req\})$\\
		\Begin{
			$FP_{failure\_type} \leftarrow failure\_probability(failure\_type)$\\
			$diff\leftarrow surv\_req(dataID)-(1-FP_{failure\_type})$\\
			\If{$neighborhood\_size > 0$}{
				\If{$diff > 0$}{
					$surv\_req(dataID)\leftarrow (diff/neighborhood\_size);$
				} \Else{
					$surv\_req(dataID)\leftarrow 0$
				}
				push $data\{dataID,surv\_req\}$ to $B$
			} \Else{
				\If{$diff > 0$}{
					push $data\{dataID,surv\_req\}$ to $Q$
				}
			}
			\If{$B \neq empty$}{
				transmit pending data items
			}
		}

		$\textbf{\emph{ScoutCheckQ}}()$\\
		\Begin{
			\For{each data item in $Q$}{
				pop $data\{dataID,surv\_req\} \leftarrow Q$\\
				\If{$surv\_req(dataID) > 0$}{
					$ScoutSendData(data\{dataID,surv\_req\})$
				}
			}
		}

		$\textbf{\emph{ScoutReceiveHello}}()$\\
		\Begin{
			$ScoutCheckQ()$\\
		}		
 
		$\textbf{\emph{ScoutReceiveData}}(data\{dataID,surv\_req\})$\\
		\Begin{
			push $data\{dataID,surv\_req\}$ to $S$\\
			\If{$surv\_req(dataID) > 0$}{
				$ScoutSendData(data\{dataID,surv\_req\})$
			}
			$ScoutCheckQ()$\\
		}

		$\textbf{\emph{ArchivistReceiveData}}(data\{dataID,surv\_req\})$\\
		\Begin{
			push $data\{dataID,surv\_req\}$ to $S$
		}
	\caption{The pseudocode for failure-adaptive and delay-tolerant replication}\label{algo:pseudo-code}
\end{algorithm}

\section{Evaluation}
\label{sec:evaluation}

In our experimental evaluations, we aim to understand: a) what improvements the proposed adaptive replication technique brings, in terms of better meeting the data survivability requirements and reducing communication costs in comparison to flooding and broadcasting techniques, under different failure models, failure rates and data types; and, b) how environmental parameters such as the size of operation affect the relative performance of the data replication techniques.

\subsection{Experimental Setup}

We used the Network Simulator $NS$--$2$~\cite{ns-2} for our evaluations, with the Shadowing Propagation Model, a path loss exponent of $3$ (typical to urban environment simulations), and a standard signal deviation of $6$ (for obstructed communication simulation).
We considered a transmission range of $250m$ and a conservative bandwidth of $11Mbps$ for the $IEEE 802.11$ protocol.
In our experiments, we used two operation area sizes: a $2Km \times 2Km$ university campus (Small Area) and a $5Km \times 5Km$ city region (Large Area).
We limited the number of types of robots to three: scouts, archivists and supervisors.
However, the archivists and supervisors are assigned the same task with respect to collecting and storing data for higher survivability, but in different types of areas.
All robots are moving according to a modified Random Waypoint Mobility model within their assigned areas; we ensure that the archivists always move forward until they switch direction when reaching the limits of their area.
We ran each experiment for $1000$ seconds of simulated time, enough for the archivists to cover their assigned connecting areas.
Our simulation results were averaged over $5$ randomly generated mobility scenarios.

Scouts produce packets with three different survivability requirements at the same rate (a packet every $3$--$7$ seconds), amounting to about $20,000$ data items of each type during the simulation time.
The only communication in these experiments is for data replication.
In the adaptive technique, robots (scouts and archivists) include in their $HELLO$ messages only the IDs of the last $10$ data items received or sent.
Even though this allows for a $HELLO$ message of realistically small size of a few $KBytes$, it also limits the amount of repository history reported as logged on the robots.
In order to compare the performance of this limited repository history ($AdLH$) with the ideal full repository history ($AdFH$) (i.e., when robots report all the repository of their locally logged data to reduce unnecessary duplicate transmissions), we ran simulations for the adaptive technique when the robots exchange a full history of the data items received or sent, while maintaining the same $HELLO$ message size to keep the network overhead constant and comparable to the replication version with limited repository history.

We applied the three failure models described in Section~\ref{sec:failure-model} with different failure rates within the range of $0\%, 10\%, 20\%, ..., 90\%$.
Furthermore, only scouts (i=1) are allowed to fail, whereas archivists (and supervisors) (i=2) are considered highly reliable and fault tolerant, thus $\rho_{1} \neq 0 \text{ and }  \rho_{2} = 0$.
Similarly, the area failures (independent or clustered) are applied only on the working areas (j=1) and not on the connecting or monitoring areas (j=2), thus $\lambda_{1} \neq 0 \text{ and } \lambda_{2} = 0$.
Consequently, given the above restrictions, we have the following typical cases for the calculation of the failure probability.
For the first failure type, when a robot has only scouts in its neighborhood, $|RT|=1$, and $FP_1 = (\rho_{1})^{a_1}$ (from eq.~\ref{eq:1}).
When the robot has scouts and at least one archivist or supervisor in its neighborhood, $|RT|=2$, and $FP_1 = (\rho_{1})^{a_1} (\rho_{2})^{a_2} = 0$ (from eq.~\ref{eq:1}).
For the second failure type, when a robot has only scouts in its neighborhood, $|AT|=1$, and $FP_2 = (\lambda_{1})^{b_1}$ (from eq.~\ref{eq:2}).
When the robot has scouts and at least one archivist or supervisor in its neighborhood, $|AT|=2$, and $FP_2 = (\lambda_{1})^{b_1} (\lambda_{2})^{b_2} = 0$ (from eq.~\ref{eq:2}).
For the third failure type, when a robot has only scouts in its neighborhood, $|AT|=1$, and $FP_3 = \lambda_{1}$ (from eq.~\ref{eq:3}).
When the robot has scouts and at least one archivist or supervisor in its neighborhood, $|AT|=2$, and $FP_3 = (\lambda_{1})(\lambda_{2}) = 0$ (from eq.~\ref{eq:3}).

We allowed an initial warm-up period of $100$ seconds before failures start.
Independent robot failures and independent area failures were uniformly distributed over time and the operation area.
However, the clustered area failures occurred simultaneously in the last $30\%$ of the simulation time, mirroring the real-case scenario of a large localized area failure (explosion, bridge collapsing, etc).
The ranges of values for the parameters used in simulations are shown in Table~\ref{tab:sim-params}.
We simulated the $5$ replication techniques presented in Section~\ref{sec:repl-surv} with $10$ different failure rates over $3$ different failure types under $5$ different mobility scenarios and $2$ operation area sizes, for a total of $1500$ simulations.

\begin{table*}
\begin{center}
\begin{tabular}{|l||l|} \hline
\textbf{Parameter}		&	\textbf{Values Used}										\\ \hline \hline
Operation Area size		&	$5Km \times 5km$, $2Km \times 2km$						\\ \hline
Number of Robots		&	$115$												\\ \hline
Working Areas			&	Size: $100m \times 100m$								\\
					&	Number of areas: $33$ (randomly placed in the operation area)	\\
					&	Robots: $3$ scouts/working area 							\\
					&	Speed Range: $1$--$5m/s$								\\ \hline
Connecting Areas		&	Size: $5Km \times 50m$, $2Km \times 50m$					\\
					&	Number of areas: $6$ ($3$ vertical, $3$ horizontal)				\\
					&	Robots: $2$ archivists per connecting area					\\
					&	Speed Range: $5$--$10m/s$								\\ \hline
Monitoring Areas		&	Size: $2.5Km \times 2.5Km$, $1Km \times 1Km$				\\
					&	Number of areas: $4$ (equally dividing the operation area)		\\
					&	Robots: $1$ supervisor per monitoring area					\\
					&	Speed Range: $10$--$15m/s$								\\ \hline
Data item size			&	$500$ Bytes per packet									\\ \hline
Data item creation period	&	$3$--$7$ seconds										\\ \hline
$HELLO$ message period	&	$8$--$12$ seconds									\\ \hline
Failure Rates			&	$0\%, 10\%, 20\%, \dots, 90\%$							\\ \hline
Survivability Requirement	&	$50\%$, $75\%$, $100\%$								\\ \hline
\end{tabular}
\caption{Simulation parameters and values used}
\label{tab:sim-params}
\end{center}
\end{table*}

\subsection{Performance Metrics}

To capture the total performance of the system at the end of the operation, we measure two cumulative performance metrics:
1)~the \textbf{Cumulative Deviation CD} which evaluates the accuracy in meeting the data survivability objectives of the operation over all data types, and
2)~the \textbf{Cumulative Replication Factor CRF} that evaluates replication redundancy and subsequently communication overhead and redundant transmissions over all data types.

Each data item of type $s$ has a Survivability Requirement $SR_{s}$, where $s\in S=\{1,2,3\}$ and $SR_s = \{100\%, 75\%, 50\%\}$.
During an experiment with a failure type $m \in M=\{1, 2, 3\}$ and failure rate $p \in P=\{0\%, 10\%, 20\%, \dots, 90\%\}$, data items of type $s$ acquire an average survivability score, $SA_{s}$ which can be lower or higher than the respective $SR_s$.
We define this average score $SA_s$ of data items of type $s$, as follows:

\[
SA_s =
\frac
{
\begin{array}{c}
\text{Distinct items of type $s$ found on surviving robots}\\
\text{at the end of the operation}
\end{array}
}
{
\begin{array}{c}
\text{Distinct items of type $s$ produced by robots}\\
\text{during the operation}
\end{array}
}
\]

Then, the \textbf{Cumulative Deviation (CD)} for a given failure type $m$ and rate $p$, over all data types $s$, is defined as follows:
\begin{equation}\label{eq:cd}
CD=
\sum_{s \in S}
|SA_{s}-SR_s|
\text{ , }
\forall m \in M
\text{ , }
\forall p \in P
\end{equation}
The ideal CD is $0$, i.e. the replication technique matches exactly the data survivability needed by all data types with their data survivability acquired over the operation.
Given that we apply three types of data with $SR_s = \{100\%, 75\%, 50\%\}$, by definition, the starting $CD$ for failure rate $p=0\%$, for all failure types and replication techniques, will be $CD=75$. 

Each data item of type $s$ can be replicated multiple times on various robots.
We define the replication factor $RF_s$ of data items of type $s$, as follows:

\[
RF_s =
\frac
{
\begin{array}{c}
\text{Number of items of type $s$ found on surviving robots}\\
\text{at the end of operation}
\end{array}
}
{
\begin{array}{c}
\text{Distinct items produced by robots}\\
\text{during the operation}
\end{array}
}
\]

Then, the \textbf{Cumulative Replication Factor (CRF)} for a given failure type $m$ and rate $p$, over all data types $s$, is defined as follows:
\begin{equation}\label{eq:crf}
CRF=
\sum_{s \in S}
RF_s
\text{ , }
\forall m \in M
\text{ , }
\forall p \in P
\end{equation}

\subsection{Results for Large Operation Area ($5Km \times 5Km$): a City Region}

\begin{figure*}[htbp]
	\centering
	\includegraphics[scale=0.65]{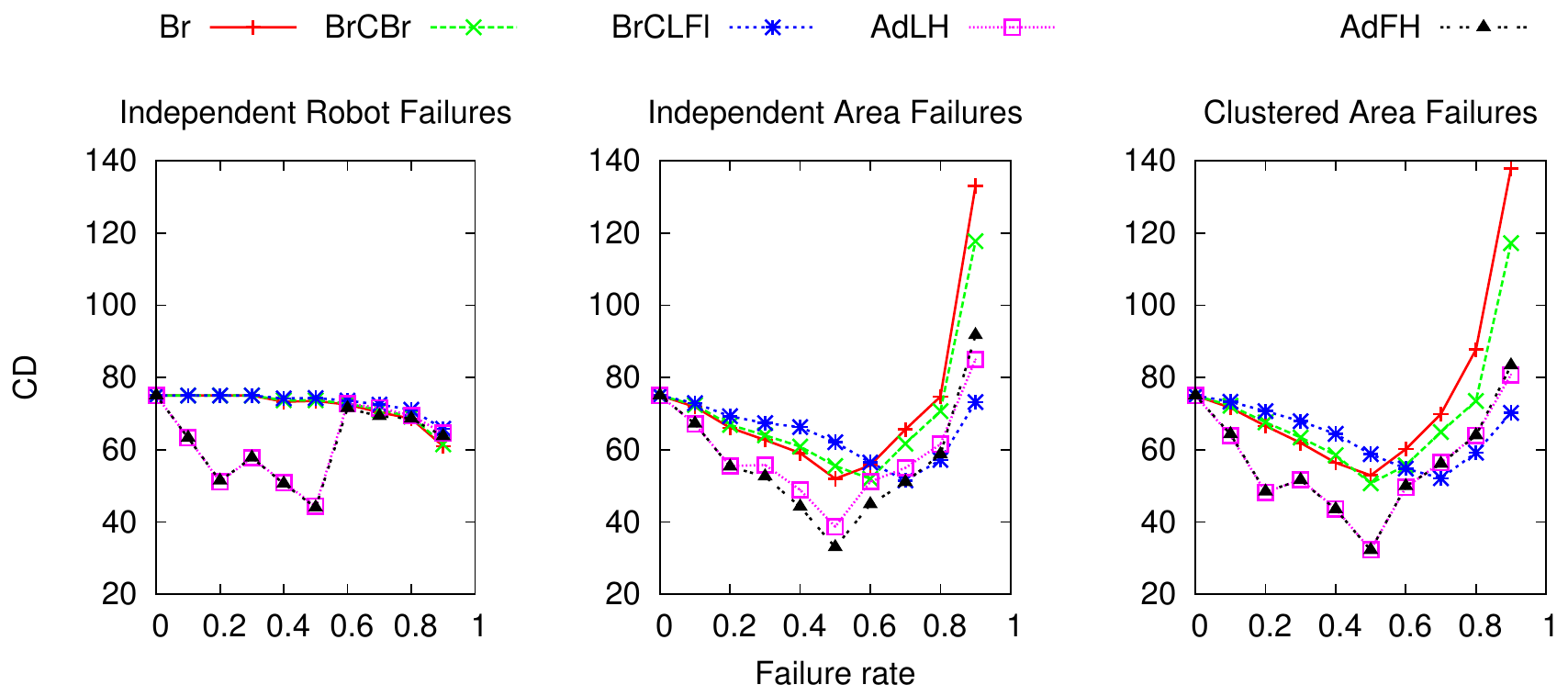}
	\caption{Large area, Cumulative Deviation ($CD$) for five different replication techniques. $Br$: Broadcast, $BrCBr$: Broadcast and Cumulative Broadcast, $BrCLFl$: Broadcast and Cumulative Limited Flooding, $AdLH$: Failure Adaptive and Delay Tolerant (limited repository history), $AdFH$: Failure Adaptive and Delay Tolerant (full repository history).
	}\label{fig:cd-large}
\end{figure*}

Figure~\ref{fig:cd-large} presents the results on cumulative deviation for the five different replication techniques, under the three failure models and different failure rates, for a large operation area of 25 square kilometers.
We notice that localized area failures lead to a higher cumulative deviation compared to independent robot failures, due to the large operation area and subsequent sparsity of the robot network.
The adaptive technique ($AdLH$ and $AdFH$) performs considerably better, adjusting the replication efforts to match the survivability requirements: for low and medium level failure rates, i.e., below $60$--$70\%$, the cumulative deviation across all types of data is almost half in comparison to the other techniques, for all three types of failures.
The broadcasting techniques ($Br$ and $BrCBr$) perform fairly well in comparison to the flooding technique ($BrCLFl$).
We decided to exclude the simple flooding from the graphs, as it performs similarly to $BrCLFl$ but with higher network overhead.

\begin{figure*}[htbp]
	\centering
	\includegraphics[scale=0.65]{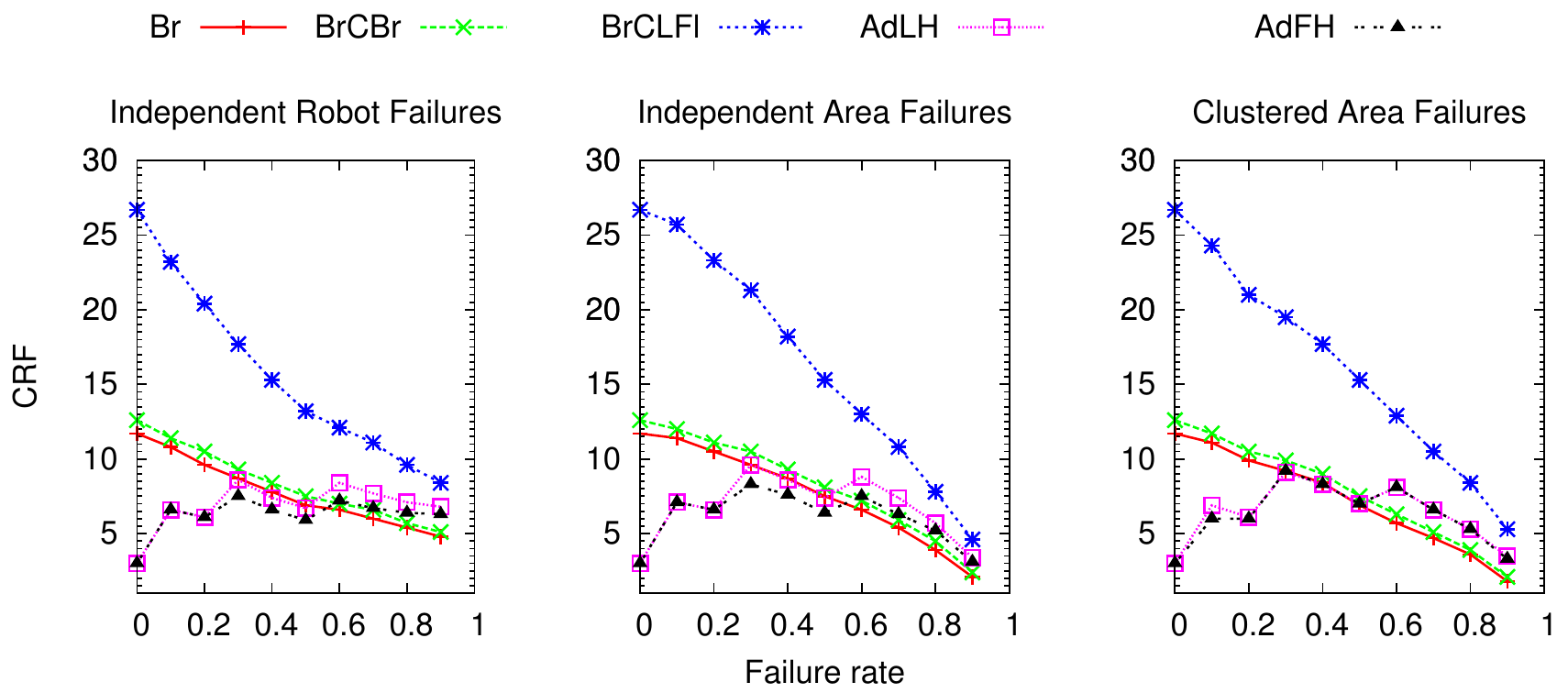}
	\caption{Large Area, Cumulative Replication Factor ($CRF$) for five different replication techniques. $Br$: Broadcast, $BrCBr$: Broadcast and Cumulative Broadcast, $BrCLFl$: Broadcast and Cumulative Limited Flooding, $AdLH$: Failure Adaptive and Delay Tolerant (limited repository history), $AdFH$: Failure Adaptive and Delay Tolerant (full repository history).
	}\label{fig:crf-large}
\end{figure*}

Figure~\ref{fig:crf-large} shows the cumulative replication factor for the five different replication techniques, under the three failure models and different failure rates for the large operation area.
We notice that the adaptive technique creates the lowest number of replicas and thus induces the lowest network overhead in the system when the anticipated failures are in low or medium levels, i.e., below $60\%$.
For high level of failure rates, i.e., above $60\%$, the adaptive technique performs similarly to broadcasting ($Br$ and $BrCBr$).
Flooding ($BrCLFl$) performs the worst: even though the technique first broadcasts and then cumulatively floods on a small radius around the transmitting scout (i.e., with a small $TTL$), the redundant replication induced by the flooding part of the method increases its cost dramatically, especially when the failure rates are at low levels, i.e., below $30\%$.
As shown in Figures~\ref{fig:cd-large} and~\ref{fig:crf-large}, the adaptive method with full history performs better than the other techniques for an additional $10$--$20\%$ of failures.
This demonstrates that the full history helps meet the survivability requirements even better than the limited history (as shown by the performance metric $CD$), while reducing communications and redundant transmissions (as shown by the performance metric $CRF$).

\subsection{Results for Small Operation Area ($2Km \times 2Km$): a University Campus}

In the small operation area setup, the monitoring and connecting areas are smaller, while the working areas remain the same size, but placed closer to each other, increasing network density.
All other parameters are preserved the same.

\begin{figure*}[htbp]
	\centering
	\includegraphics[scale=0.65]{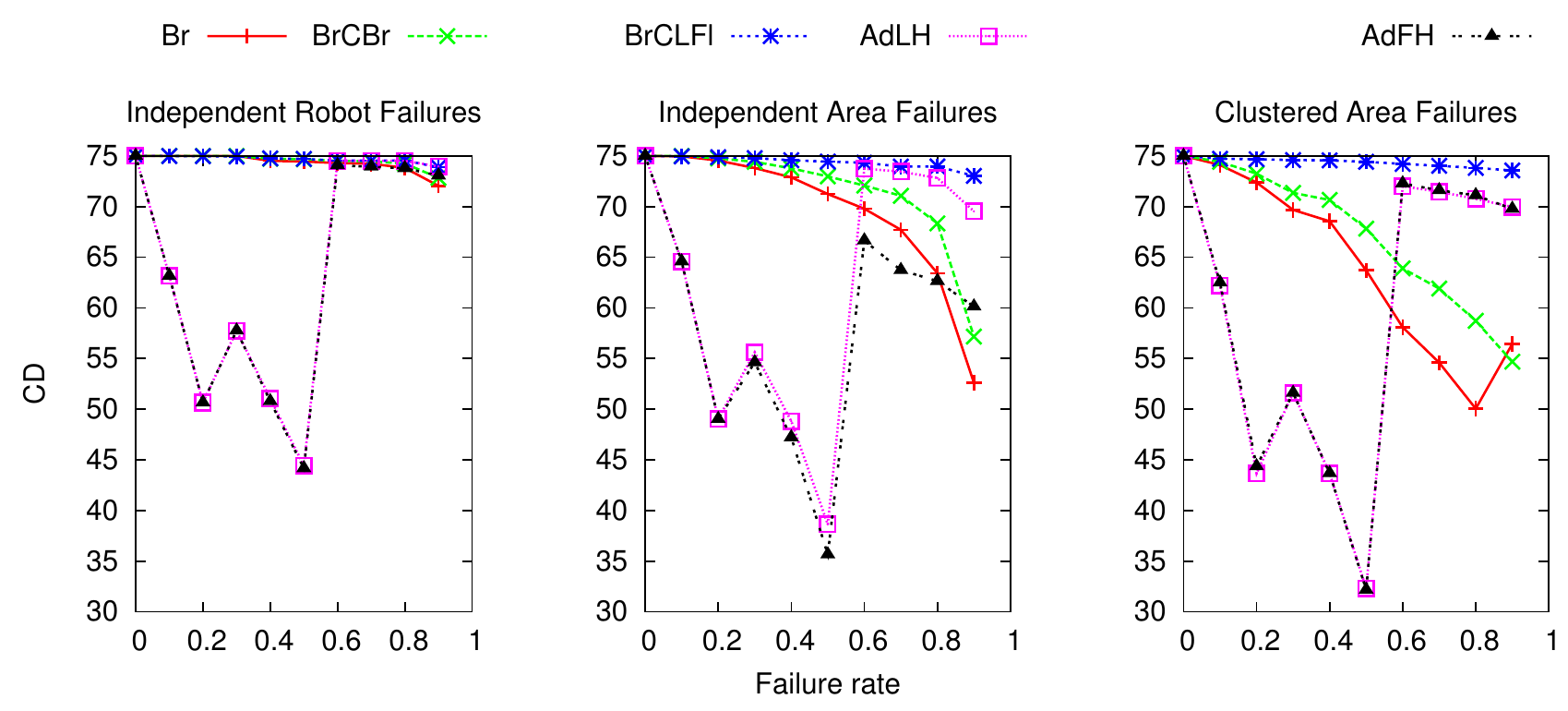}
	\caption{
	Small area, Cumulative Deviation ($CD$) for five different replication techniques. $Br$: Broadcast, $BrCBr$: Broadcast and Cumulative Broadcast, $BrCLFl$: Broadcast and Cumulative Limited Flooding, $AdLH$: Failure Adaptive and Delay Tolerant (limited repository history), $AdFH$: Failure Adaptive and Delay Tolerant (full repository history).
	}\label{fig:cd-small}
\end{figure*}

Figure~\ref{fig:cd-small} shows the cumulative deviation for the five different replication techniques, under the three failure models and different failure rates, for a small operation area of 4 square kilometers.
In contrast to the large operation area, we observe similar performance patterns for all three types of failures due to a significant increase in network density and thus reachability between robots.
The adaptive technique exhibits half the cumulative deviation of the other techniques, for low and medium level failure rates, i.e., up to $60\%$.
As expected, in this smaller area of operation (about $6$ times smaller than the large area), flooding is not a viable option as it deviates the most from the data survivability requirements.
Nevertheless, broadcasting performs well, especially in high failure rates.

\begin{figure*}[htbp]
	\centering
	\includegraphics[scale=0.65]{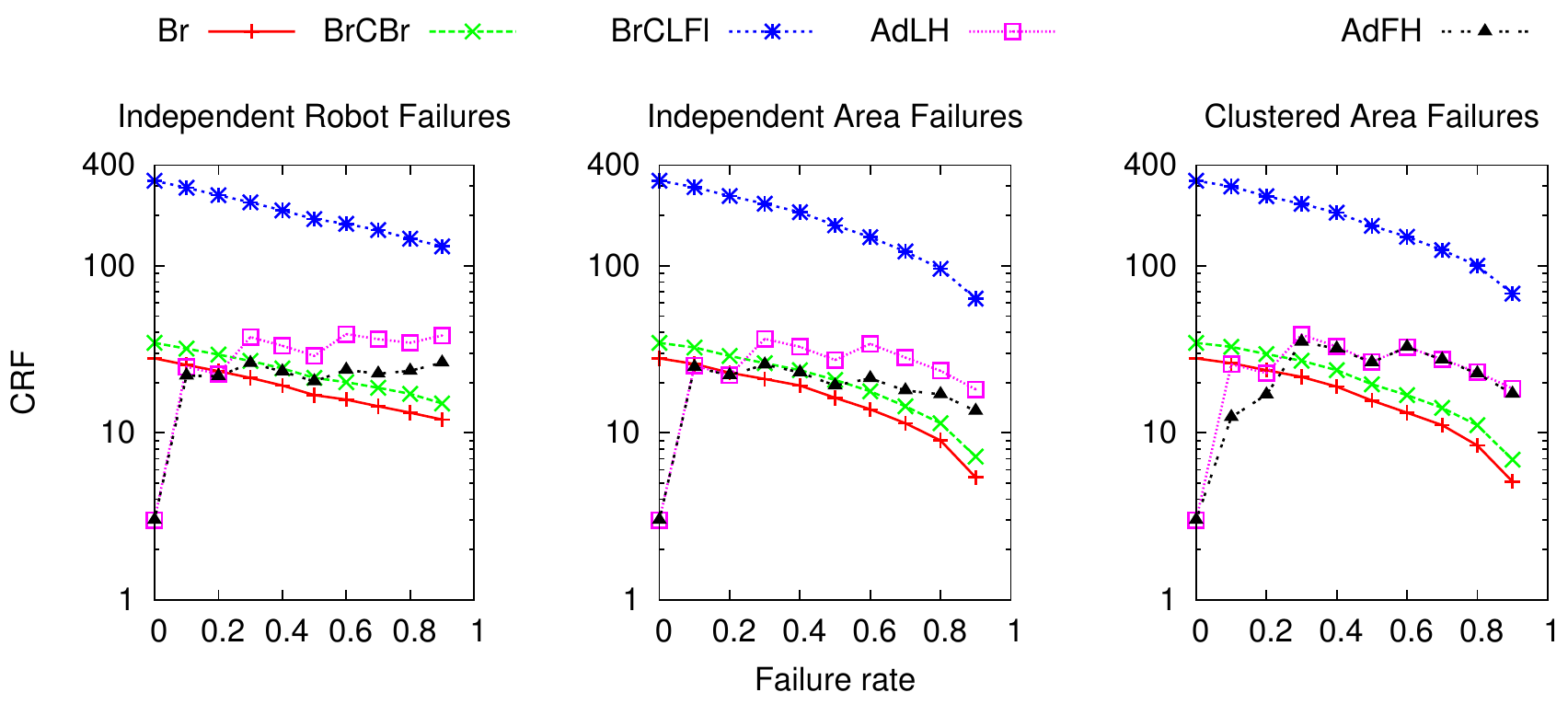}
	\caption{
	Small Area, Cumulative Replication Factor (CRF) for 5 different replication techniques. Notice the log scale in y-axis. $Br$: Broadcast, $BrCBr$: Broadcast and Cumulative Broadcast, $BrCLFl$: Broadcast and Cumulative Limited Flooding, $AdLH$: Failure Adaptive and Delay Tolerant (limited repository history), $AdFH$: Failure Adaptive and Delay Tolerant (full repository history).
	}\label{fig:crf-small}
\end{figure*}

Figure~\ref{fig:crf-small} shows the cumulative replication factor for the five different replication techniques, under the three failure models and different failure rates for the small operation area.
We notice that the adaptive technique does not perform as well as in the large area due to the overhearing effect:
1)~The transmitting robot calculates a possible failure probability for its neighborhood and sends an item.
2)~Robots in transmission range that were not accounted in the failure probability calculation (e.g. their $HELLO$ message was dropped, or came within reach afterwards), receive the item, store it locally and contribute to its acquired survivability, but also to its replication factor.
3)~However, the transmitting robot calculated a higher failure probability, thus the needed survivability from the receiving robots is set to a higher level and more replicas are potentially applied to reach the survivability requirement.
Flooding performs the worst, inducing high communication and redundant replication (more than an order of magnitude in comparison to the large area).

\subsection{Summary of Experimental Results}

Our simulations of urban search and rescue scenarios within disaster environments show how the performance of various replication techniques depends on the operation area size and the type and rate of failures.
Especially in a large operation area, data survivability requirements are more difficult to meet when area failures (independent or clustered) are considered, in comparison to the independent robot failures.
Therefore, the replication technique used must be more adaptive and aggressive, to provide higher guarantees of meeting the data survivability requirements.
On the other hand, a smaller operation area leads to similar performance patterns across all failure types, due to higher network density.
Thus, in such an operational setup with high network density it is less critical to specify exactly the failure type anticipated during the mission.

However, from the results in both operation areas, the failure rate is the decisive factor for the performance in data survivability, and can be discretized into three ranges: low ($0\%~-~30\%$), medium ($30\%~-~60\%$) and high ($60\%~-~90\%$).
Consequently, the human coordinators can choose among these three levels instead of an exact failure rate.
Furthermore, the failure rate level can be assumed high in the beginning of the mission and gradually reduced to lower levels as the coordinators have more empirical evidence at hand.

The amount of data that survives in the network depends on how aggressive the technique is to replicate in other than the producer's area, compensating for the replicas to be lost due to local robot failures.
The adaptive technique best matches the survivability requirements of data for up to medium level of anticipated failure rates (up to $60\%$), while reducing the number of replicas in the network (compared to flooding and broadcasting techniques), thus reducing communications overhead, battery and storage usage.

\section{Related Work}
\label{sec:rel-work}

The major differences between our work and previous studies is along three coordinates: (i) heterogeneity in nodes, types of failures, and data requirements; (ii)~large-scale network-level simulations, and (iii)~ data replication in the context of a heterogeneous environment.

Studies that have considered node heterogeneity assume that some nodes are more stable and more resourceful than others.
CLEAR~\cite{mondal06clear} deploys a super-peer architecture that exploits relatively stable peers having maximum remaining battery power and processing capacity among their regional neighbors to determine a near-optimal reallocation period based on mobile host schedules.
In~\cite{abdulla07corecopies}, resourceful nodes serve as cores to enable core-aided routing. 
Replication schemes examined include ``copy-to-core'', where both regular nodes and core nodes are carriers of messages to the destination, and ``dump-to-core'', where the regular nodes delete the messages, leaving the cores to deal with the delivery.
Such nodes, due to their extended resources, acquire a similar role to the \textit{archivists} and \textit{supervisors} presented in this work.
However, unlike in our work, no differentiation was made with respect to their mobility or failures exhibited and how they affect the replication effort.

Node failures in robotic networks or MANETs are usually considered independent and homogeneous.
A deviation from the independent node failure model was studied in Replic8~\cite{kotsovinos05replic8} which considered clustered node failures that affect multiple nodes at the same time.
Replic8 performs location-aware file replication and tackles localized network failures by storing replicas at faraway network locations and achieves high availability with fewer replicas as compared to replicating at random locations.
Similarly, our work applies failure models that affect multiple nodes at the same time.
However, we distinguish between independent area failures, where same team robots working in an area can fail at the same time independently of neighboring teams, and clustered area failures, where multiple neighboring areas can fail at the same time.

Heterogeneity in data has been used in prioritized epidemic routing (PREP)~\cite{ramanathan07prep}, were bundles of data are prioritized based on cost to destination, source, and expiration time.
Costs are derived from per-link ``average availability'' information that is disseminated in an epidemic manner.
PREP maintains a gradient of replication density that decreases with increasing distance from destination.
In our work, we also assume that data are replicated based on different requirements or priorities.
However, these requirements are due to the significance  to survive the mission, as assumed by the human coordinators and not due to their producer, their destination or associated lifetime.
Our adaptive replication technique adjusts (i.e., increases) the survivability acquired by the data as they are replicated across new robots, until their accumulated survivability meets their original requirements set by the coordinators.

Research in MANETs traditionally uses simulations at network level with a focus on scalability, but considers nodes homogeneous and failures independent; in robotics research the emphasis is more on studying the features of a particular node or improving the collaboration between nodes in relatively small teams, and thus traditionally focuses less on scale or low-level communication aspects.
Our study bridges this gap by leveraging MANET traditional tools (low-level network simulators, in this particular case NS-2) in a large-scale environment populated by heterogeneous mobile nodes with differentiated characteristics defined by robotics scenarios such as available robot resources and sensor data production, robot mobility and anticipated failures.

Data replication has been studied extensively in MANETs.
Three replication techniques to improve data accessibility are presented in~\cite{hara02replicas1,hara03replicas2} where data access frequency is used to place replicas, eliminate duplications among neighboring hosts, and share replicas among stable groups of hosts.
In~\cite{chen01lookupreplicas}, nodes exchange availability information by broadcasting advertisement messages to decrease redundancy and create a data lookup table within their group.
Group partitioning prediction allows replication of data before partitioning occurs.
In~\cite{jing04replicas} Jing et al propose an algorithm that takes into account the nodes' motion and tries to minimize the communication cost of data access.
The algorithm attempts to adapt dynamically the replica allocation scheme to a local optimal scheme.
A theoretical approach that quantifies the effect of data replication on availability in MANETs is presented in~\cite{gianuzzi04effectiveness}.
In~\cite{zheng04stable} Zheng et al use read and write statistics to define the level of replication in a neighborhood of nodes.
In~\cite{liangzhong06caching}, the peers define data cache policies based on requests to access particular data.
In contrast to these studies on data replication for availability and accessibility in real time, we focus on the problem of data survivability, which is data durability for the duration of the operation, given different data survivability requirements and node failures. 

Employing lessons from~\cite{frew09networking} on UAVs, our adaptive method uses opportunistic communication (similar to delay tolerant networking) between all types of robots (UGVs, USVs and UAVs), to improve the survivability of data in the system in face of robot and network failures.
Several studies (e.g.,~\cite{abdulla07corecopies, vahdat00epidemic, goundan08broadcast, harras09flooding, ignacio08disrupted, ramanathan07prep, spyropoulos05sprayandwait}) focus on challenged networks and employ delay-tolerant techniques for data replication.
A disruption resilient content dissemination approach for MANETs~\cite{ignacio08disrupted} exploits the in-network storage and hop-by-hop dissemination of named information objects.
A broadcast scheme for delay tolerant networks is presented in~\cite{goundan08broadcast}, where a node transmits a message only when at least one node in range does not have the particular message.
Our adaptive technique applies a similar approach: upon receiving a $HELLO$ message from an archivist, a scout reactively transmits the history of data since the last $HELLO$ from an archivist.
With ``Spray and Wait''~\cite{spyropoulos05sprayandwait}, a node ``sprays'' a number of copies into the network and then ``waits'' until one of the sprayed nodes meets the destination.
In~\cite{vahdat00epidemic}, epidemic routing is used as a controlled flooding technique, where a pair of nodes exchange missing packets when into contact.
Given enough storage space, epidemic routing can be used to reliably disseminate data across the network.
An evaluation of different controlled message flooding schemes over disconnected sparse mobile networks is presented in~\cite{harras09flooding}.
Similarly to these efforts, our adaptive technique also uses a delay-tolerant mechanism.
However, our ultimate goal is to increase the survivability of data with the least network overhead induced in the system.
Additionally, instead of high density mobile networks comprised of homogeneous nodes producing one type of data, we consider scenarios with low, non-uniform node density and heterogeneous nodes which continuously produce data of different types and with different survivability requirements.

Significant research has been done to provide realistic mobility models for MANETs.
Of particular relevance for our work are group mobility models, such as the ones presented in~\cite{cano04groupmobility},~\cite{hong99groupmobility},~\cite{ng05groupmobility} and~\cite{wang02groupmobility}.
These can be used to model the robot mobility inside the areas defined in this paper, instead of the Random Waypoint Model used.
Also, a model for realistic representation of the movement of civil protection units in a disaster area scenario was presented in~\cite{aschenbruck07mobilitydisaster}.
This study focused on a small scale operation area such as a collapsed building (hundreds of square meters) with a high network density and high reachability between nodes.
Instead, our study considers large operation areas (tens of square kilometers) that include several small working areas (e.g., collapsed buildings) and nodes forming a fairly sparse network.
Moreover, node failures during the operation reduce further network density and challenge data survivability.

\section{Conclusions}
\label{sec:conclusions}

Robots typically have more resources than, for example, smart phones or sensors, but are more prone to fail.
At the same time, the data they collect when deployed in urban disaster environments can be vital for search and rescue efforts.
This paper studies the problem of data survivability in mobile robot networks by acknowledging that, unlike nodes whose mobility patterns cannot be externally controlled or accurately predicted (e.g., cell phones), more information can be available in robot networks and can be used for failure-resistant data collection in disaster scenarios.
Our work bridges the fields of MANETs and Robotics by studying the data survivability using large-scale, network-level robotic simulations, while leveraging human assessment of the environment and robot heterogeneity.

We proposed a delay-tolerant failure-adaptive replication technique that takes into account the following mission-specific and environmental parameters.
First, data collected during a mission have different survivability requirements based on their importance to the human crews.
This mission-specific parameter can be easily estimated function of the existing sensors for collecting data and the deployment scenario.
For example, in a nuclear power plant accident, video footage from the on-board robot cameras while inspecting the structural integrity of the site can be assigned a survivability requirement of $100\%$, radioactivity level readings can be assigned a survivability requirement of $75\%$ (by default, rescue crews enter the site with radioactive-resistant suits), whereas temperature and humidity readings can be assigned a survivability requirement of $25\%$.

Second, robots are assigned different tasks based on their hardware characteristics and mission.
For example, robot scouts can work as teams to collect data and complete a common task.
On the other hand, robot archivists can follow paths that connect scout teams, for better control, wireless connectivity and data backups.
Depending on the robot type and assigned task, robots can exhibit different failure rates, which makes robot heterogeneity an important mission-specific parameter.

Third, robotic failures of various types and rates lead to data loss and greatly affect the success of a mission.
We classified these failures into three categories: independent robot failures, independent area failures and clustered area failures.
Human operators could estimate the type of robot failure expected during a mission based on environmental conditions and robot types used.
For example, urban architecture information such as the placement of gas pipes or topology could lead to explosions or, respectively, flooding, and thus to clustered area failures, whereas a preliminary or conservative estimation of the state of a building or bridge structure could determine area failures.
Finally, empirical estimation of robot reliability based on hardware specifications and task could suggest independent robot failures.

The rate of failures is difficult to estimate accurately, especially for clustered and independent area failures.
We acknowledge that in many cases the value of the anticipated failure rate may be just an educated guess.
However, even in such cases, instead of the exact rate, an average level of failure rate could be used.
Our results indicate three general levels of failure rate that can inform adaptive replication decisions:
given a pessimistic estimation of the situation, the operation coordinators can set the failure rate to a high level, and later on reduce it to medium or low levels based on empirical evidence.

These parameters are used by our replication technique to find the right tradeoff between replication (and thus resource consumption, such as battery, communication volume and storage) and data survivability.
Our technique utilizes a distributed mechanism to estimate the probability of a neighborhood to fail.
Each robot uses this estimation to adapt the replication for each type of data based on the failures anticipated in its neighborhood, which is a strong indicator of the survivability that the data can acquire during the mission.
As a future extension, this mechanism could also be used by each robot to update its failure rate at real time during the mission, instead of being set by the coordinators.
Thus, a robot could start at a default failure rate and by updating this rate based on feedback from the observed failure probability of its neighborhood, it could gradually converge to a stable and more accurate failure rate.
This situation would lead in the beginning to higher communication costs than necessary, yet still lower overall than the baseline solutions with which we compared our adaptive technique.

For our experimental study we proposed and used novel frameworks for generating realistic mobility and failure scenarios for mobile robot networks when deployed for search and rescue missions in urban disaster scenarios.
These frameworks were adapted for the Network Simulator NS-2 and allowed us to experiment with different sizes and types of areas, for different numbers and types of robots assigned in each area, with variable mobility pattern for each robot type, and finally different failure models: independent robot failures, independent area failures and clustered area failures.
Extensive network-level simulations demonstrated that our adaptive replication technique allows a mission to withstand failure rates of up to 60\% of the robots for all three failure models examined, while resulting in better data survivability than flooding and broadcasting-based techniques and without inducing higher communication costs.

Deployment of ad-hoc collaborative robot teams is currently done at a small scale and in controlled environments.
The reliability of robots and survivability of their data in the presence of environmental hazards need more studies through realistic simulations to allow the emergence of large-scale self-coordinating mobile robot networks.
Our study contributes to more realistic modeling of robot placement, mobility, failures and survivability of their data in urban disaster environments.

\section*{acknowledgements}
This research was partially supported by grant ARI W74V8H-05-C-0052 and by the National Science Foundation under Grants No. CNS-0454081, IIS-0534520, CNS-0520033 and CNS-0520123. Any opinions, findings, and conclusions or recommendations expressed in this material are those of the authors and do not necessarily reflect the views of the sponsors.

\bibliographystyle{abbrv}
\bibliography{refs}

\begin{thebibliography}{10}

\bibitem{abdulla07corecopies}
M.~Abdulla and R.~Simon.
\newblock Analysis of core-assisted routing in opportunistic networks.
\newblock In {\em 15th International Symposium on Modeling, Analysis, and
  Simulation of Computer and Telecommunication Systems}, pages 387--394,
  Istanbul, Turkey, October 24--26 2007.

\bibitem{aschenbruck07mobilitydisaster}
N.~Aschenbruck, E.~Gerhards-Padilla, M.~Gerharz, M.~Frank, and P.~Martini.
\newblock Modeling mobility in disaster area scenarios.
\newblock In {\em 10th ACM Symposium on modeling, analysis, and simulation of
  wireless and mobile systems}, pages 4--12, Chania, Crete Island, Greece,
  October 22--26 2007.

\bibitem{birk11mosaicking}
A.~Birk, B.~Wiggerich, H.~Bülow, M.~Pfingsthorn, and S.~Schwertfeger.
\newblock Safety, security, and rescue missions with an unmanned aerial vehicle
  ({UAV}).
\newblock {\em Journal of Intelligent \& Robotic Systems}, pages 1--20, 2011.

\bibitem{bouabdallah10palm-sizeheli}
S.~Bouabdallah, C.~Bermes, S.~Grzonka, C.~Gimkiewicz, A.~Brenzikofer, R.~Hahn,
  D.~Schafroth, G.~Grisetti, W.~Burgard, and R.~Siegwart.
\newblock Towards palm-size autonomous helicopters.
\newblock {\em Journal of Intelligent \& Robotic Systems}, 61:445--471, 2010.

\bibitem{burkle10uav-swarms}
A.~Burkle, F.~Segor, and M.~Kollmann.
\newblock Towards autonomous micro {UAV} swarms.
\newblock {\em Journal of Intelligent \& Robotic Systems}, 61:339--353, 2010.

\bibitem{cano04groupmobility}
J.-C. Cano, P.~Manzoni, and M.~Sanchez.
\newblock Evaluating the impact of group mobility on the performance of mobile
  ad hoc networks.
\newblock In {\em IEEE International Conference on Communications}, pages
  4039--4043, Paris, France, June 20--24 2004.

\bibitem{carson03reliability}
J.~Carlson and R.~Murphy.
\newblock Reliability analysis of mobile robots.
\newblock In {\em IEEE International Conference on Robotics and Automation},
  pages 274--281, Taipei, Taiwan, September 14-19 2003.

\bibitem{carson05failures}
J.~Carlson and R.~Murphy.
\newblock How {UGV}s physically fail in the field.
\newblock {\em IEEE Transactions on Robotics}, 3:423--437, 2005.

\bibitem{carson04failures}
J.~Carlson, R.~Murphy, and A.~Nelson.
\newblock Follow-up analysis of mobile robot failures.
\newblock In {\em IEEE International Conference on Robotics and Automation},
  pages 4987--4994, New Orleans, LA, USA, April 26-- May 1 2004.

\bibitem{chen01lookupreplicas}
K.~Chen and K.~Nahrstedt.
\newblock An integrated data lookup and replication scheme in mobile ad hoc
  networks.
\newblock {\em Society of Photo-Optical Instrumentation Engineers}, 4534:1--8,
  2001.

\bibitem{crasar11crasar}
CRASAR.
\newblock Center for {R}obot-{A}ssisted {S}earch and {R}escue.
\newblock Online, 2012, 2012.

\bibitem{dudenhoeffer00micro-robot-formation}
D.~D. Dudenhoeffer and M.~P. Jones.
\newblock A formation behavior for large-scale micro-robot force deployment.
\newblock In {\em Proceedings of the 32nd Conference on {W}inter {S}imulation},
  pages 972--982, Orlando, Florida, 2000.

\bibitem{frew09networking}
E.~Frew and T.~Brown.
\newblock Networking issues for small unmanned aircraft systems.
\newblock {\em Journal of Intelligent \& Robotic Systems}, 54:21--37, 2008.

\bibitem{gianuzzi04effectiveness}
V.~Gianuzzi.
\newblock Data replication effectiveness in mobile ad-hoc networks.
\newblock In {\em 1st ACM International Workshop on performance evaluation of
  wireless ad hoc, sensor, and ubiquitous networks}, pages 17--22, Venezia,
  Italy, October 7 2004.

\bibitem{goundan08broadcast}
A.~Goundan, E.~Coe, and C.~Raghavendra.
\newblock Efficient broadcasting in delay tolerant networks.
\newblock In {\em IEEE Global Telecommunications Conference}, pages 1--5, New
  Orleans, LA, USA, November 30--December 4 2008.

\bibitem{hara02replicas1}
T.~Hara.
\newblock Replica allocation in ad hoc networks with periodic data update.
\newblock In {\em 3rd IEEE International Conference on Mobile Data Management},
  pages 79--86, Singapore, January 8--11 2002.

\bibitem{hara03replicas2}
T.~Hara.
\newblock Replica allocation methods in ad hoc networks with data update.
\newblock {\em Mobile Networks and Applications}, 8:343--354, 2003.

\bibitem{harras09flooding}
K.~A. Harras and K.~C. Almeroth.
\newblock Controlled flooding in disconnected sparse mobile networks.
\newblock {\em Wireless Communications and Mobile Computing}, 9:21--33, 2009.

\bibitem{hong99groupmobility}
X.~Hong, M.~Gerla, G.~Pei, and C.-C. Chiang.
\newblock A group mobility model for ad hoc wireless networks.
\newblock In {\em 2nd ACM International workshop on modeling, analysis and
  simulation of wireless and mobile systems}, pages 53--60, Seattle, WA, August
  20 1999.

\bibitem{spectrum11japansurvivors}
IEEE-Spectrum.
\newblock Japan earthquake robots help search for survivors.
\newblock Online, 2012, 2011.

\bibitem{spectrum11japanreactors2}
IEEE-Spectrum.
\newblock Robotic aerial vehicle at fukushima reactors.
\newblock Online, 2012, 2011.

\bibitem{spectrum11japanreactors1}
IEEE-Spectrum.
\newblock Robots enter fukushima reactors and detect high radiation.
\newblock Online, 2012, 2011.

\bibitem{ignacio08disrupted}
S.~Ignacio and J.~Garcia-Luna-Aceves.
\newblock Robust content dissemination in disrupted environments.
\newblock In {\em 3rd ACM Workshop on challenged networks}, pages 3--10, San
  Francisco, CA, USA, September 14--19 2008.

\bibitem{jing04replicas}
Z.~Jing, W.~Yijie, L.~Xicheng, and Y.~Kan.
\newblock A dynamic adaptive replica allocation algorithm in mobile ad hoc
  networks.
\newblock In {\em 2nd IEEE Conference on Pervasive Computing and Communications
  Workshops}, pages 65--69, Orlando, FL, USA, March 14--17 2004.

\bibitem{kleiner11mapping}
A.~Kleiner and C.~Dornhege.
\newblock Mapping for the support of first responders in critical domains.
\newblock {\em Journal of Intelligent \& Robotic Systems}, pages 1--25, 2011.

\bibitem{kotsovinos05replic8}
E.~Kotsovinos and D.~McIlwraith.
\newblock Replic8: Location-aware data replication for high availability in
  ubiquitous environments.
\newblock In {\em 3rd International Conference in Wired/Wireless Internet
  Communications}, volume 3510, pages 32--41, Xanthi, Greece, May 11--13 2005.
  Lecture Notes in Computer Science, Springer.

\bibitem{liangzhong06caching}
Y.~Liangzhong and C.~Guohong.
\newblock Supporting cooperative caching in ad hoc networks.
\newblock {\em IEEE Transactions on Mobile Computing}, 5:77--89, 2006.

\bibitem{maza10multi-uav}
I.~Maza, F.~Caballero, J.~Capitan, J.~Martinez-de Dios, and A.~Ollero.
\newblock Experimental results in multi-{UAV} coordination for disaster
  management and civil security applications.
\newblock {\em Journal of Intelligent \& Robotic Systems}, 61:563--585, 2010.

\bibitem{mondal06clear}
A.~Mondal, S.~Madria, and M.~Kitsuregawa.
\newblock {CLEAR}: An efficient context and location-based dynamic replication
  scheme for mobile-p2p networks.
\newblock In {\em 17th International Conference on Database and Expert Systems
  Applications}, pages 399--408, Krakow, Poland, September 4--8 2006.

\bibitem{murphy04security}
R.~Murphy.
\newblock Rescue robotics for homeland security.
\newblock {\em Communication ACM}, 47:66--68, 2004.

\bibitem{murphy08mine}
R.~Murphy, J.~Kravitz, K.~Peligren, J.~Milward, and J.~Stanway.
\newblock Preliminary report: rescue robot at {C}randall {C}anyon, {U}tah, mine
  disaster.
\newblock In {\em IEEE International Conference on Robotics and Automation},
  pages 2205--2206, Pasadena, CA, USA, May 19-- May 23 2008.

\bibitem{murphy11ike}
R.~Murphy, E.~Steimle, M.~Hall, M.~Lindemuth, D.~Trejo, S.~Hurlebaus,
  Z.~Medina-Cetina, and D.~Slocum.
\newblock Robot-assisted bridge inspection.
\newblock {\em Journal of Intelligent \& Robotic Systems}, pages 1--19, 2011.

\bibitem{ng05groupmobility}
J.~Ng and Y.~Zhang.
\newblock A mobility model with group partitioning for wireless ad hoc
  networks.
\newblock In {\em 3rd International Conference on Information Technology and
  Applications}, pages 289--294, Sydney, Australia, July 4--7 2005.

\bibitem{ns-2}
NS-2.
\newblock Network {S}imulator--2.
\newblock Online, 2012, 2012.

\bibitem{murphy01wtc}
NSF.
\newblock At {WTC} search, graduate students deploy shoebox-sized robots.
\newblock Online, 2012, 2001.

\bibitem{murphy05katrina1}
NSF.
\newblock Small, unmanned aircraft search for survivors in {K}atrina wreckage.
\newblock Online, 2012, 2005.

\bibitem{ramanathan07prep}
R.~Ramanathan, R.~Hansen, P.~Basu, R.~Rosales-Hain, and R.~Krishnan.
\newblock Prioritized epidemic routing for opportunistic networks.
\newblock In {\em 1st International MobiSys Workshop on mobile opportunistic
  networking}, pages 62--66, San Juan, PR, USA, June 11--14 2007.

\bibitem{robotnet11competition}
RobotsNet.
\newblock Robot competitions.
\newblock Online, 2012, 2011.

\bibitem{spyropoulos05sprayandwait}
T.~Spyropoulos, K.~Psounis, and S.~Cauligi.
\newblock Spray and wait: an efficient routing scheme for intermittently
  connected mobile networks.
\newblock In {\em SIGCOMM Workshop on delay-tolerant networking}, pages
  252--259, Philadelphia, PA, USA, August 26 2005.

\bibitem{sugiyama05collaboration}
H.~Sugiyama, T.~Tsujioka, and M.~Murata.
\newblock Collaborative movement of rescue robots for reliable and effective
  networking in disaster area.
\newblock In {\em 1st IEEE International Conference on Collaborative Computing:
  Networking, Applications and Worksharing}, San Jose, CA, USA, December 19--21
  2005.

\bibitem{sze-yao99broadcaststorm}
N.~Sze-Yao, T.~Yu-Chee, C.~Yuh-Shyan, and S.~Jang-Ping.
\newblock The broadcast storm problem in a mobile ad hoc network.
\newblock In {\em 5th IEEE International Conference on Mobile Computing and
  Networking}, pages 151--162, Seattle, WA, USA, August 15--19 1999.

\bibitem{theeagle11robotstorescue}
Theeagle.
\newblock Robots to the rescue: Machines built for dangerous situations put to
  test.
\newblock Online, 2012, 2011.

\bibitem{vahdat00epidemic}
A.~Vahdat and D.~Becker.
\newblock Epidemic routing for partially-connected ad hoc networks.
\newblock In {\em Duke University, Technical Report CS-200006}, November
  30--December 4 2000.

\bibitem{vasilescu05muling}
I.~Vasilescu, K.~Kotay, D.~Rus, M.~Dunbabin, and P.~Corke.
\newblock Data collection, storage, and retrieval with an underwater sensor
  network.
\newblock In {\em 3rd International Conference on Embedded Networked Sensor
  Systems}, pages 154--165, San Diego, California, USA, November 2--4 2005.

\bibitem{wang02groupmobility}
K.~Wang and B.~Li.
\newblock Group mobility and partition prediction in wireless ad-hoc networks.
\newblock In {\em IEEE International Conference on Communications}, pages
  1017--1021, New York, NY, USA, May 2 2002.

\bibitem{haiti10earthquake}
Wikipedia.
\newblock 2010 {H}aiti {E}arthquake.
\newblock Online, 2012, 2011.

\bibitem{japan11earthquake}
Wikipedia.
\newblock 2011 {T}$\overline{o}$hoku {E}arthquake and {T}sunami.
\newblock Online, 2012, 2011.

\bibitem{wood08flyrobot}
R.~Wood.
\newblock Fly, robot fly.
\newblock {\em IEEE Spectrum}, pages 25--29, March 2008.

\bibitem{zheng04stable}
J.~Zheng, J.~Su, K.~Yang, and Y.~Wang.
\newblock Stable neighbor based adaptive replica allocation in mobile ad hoc
  networks.
\newblock In {\em International Conference on Computational Science}, volume
  LNCS 3036, pages 373--380, Krakow, Poland, June 6--9 2004.

\end{thebibliography}

\end{document}